\documentclass[runningheads]{llncs}
\usepackage{hyperref}
\usepackage{multirow}
\usepackage{graphicx}
\newcommand{\repeatthanks}{\textsuperscript{\thefootnote}}

\begin{document}

\title{ImpactCite: An XLNet-based method for Citation Impact Analysis}
\titlerunning{ImpactCite: An XLNet-based method for Citation Impact Analysis}
%\author{First Author\inst{1}\orcidID{0000-1111-2222-3333} \and
%Second Author\inst{2,3}\orcidID{1111-2222-3333-4444} \and
%Third Author\inst{3}\orcidID{2222--3333-4444-5555}}

\author{Dominique Mercier\thanks{Equal contribution}\inst{1,2} \and
Syed Tahseen Raza Rizvi\repeatthanks\inst{1,2} \and
Vikas Rajashekar\inst{2} \and
Andreas Dengel\inst{1,2} \and
Sheraz Ahmed\inst{2}}
\authorrunning{Mercier et al.}
\institute{Technische Universität Kaiserslautern, 67663 Kaiserslautern, Germany \\
\email{\{mercier,srizvi\}@rhrk.uni-kl.de}\\ \and
German Research Center for Artificial Intelligence (DFKI), Kaiserslautern, Germany\\
\email{\{dominique.mercier, syed\_tahseen\_raza.rizvi, vikas.rajashekar, andreas.dengel, sheraz.ahmed\}@dfki.de}}
\maketitle              % typeset the header of the contribution
\begin{abstract}

Citations play a vital role in understanding the impact of scientific literature. Generally, citations are analyzed quantitatively whereas qualitative analysis of citations can reveal deeper insights into the impact of a scientific artifact in the community. Therefore, citation impact analysis (which includes sentiment and intent classification) enables us to quantify the quality of the citations which can eventually assist us in the estimation of ranking and impact. The contribution of this paper is two-fold. First, we benchmark the well-known language models like BERT and ALBERT along with several popular networks for both tasks of sentiment and intent classification. Second, we provide ImpactCite, which is XLNet-based method for citation impact analysis. All evaluations are performed on a set of publicly available citation analysis datasets. Evaluation results reveal that ImpactCite achieves a new state-of-the-art performance for both citation intent and sentiment classification by outperforming the existing approaches by 3.44\% and 1.33\% in F1-score. Therefore, we emphasize ImpactCite (XLNet-based solution) for both tasks to better understand the impact of a citation. Additional efforts have been performed to come up with CSC-Clean corpus, which is a clean and reliable dataset for citation sentiment classification.

%intent current state-of-the-art SciBert: 	85.49 -- ours 88.93
%sentiment current state-of-the-art SVM: 76.4 --> ours 77.73

\end{abstract}

\section{Introduction}
Scientific publications play an important role in the development of a community. An exponential increase in scientific literature has posed a challenge of evaluating the impact of a publication in a given scientific community. Citations majorly contribute towards the eminence of an author as well as the impact of their publications in a society. However, counting citations serves as a quantitative metric and therefore does not provide qualitative insights to the citations. In order to get a qualitative insight, the sentiment of a given citation is identified which refers to the opinion of the citing author about the cited literature. There exist generally three types of sentiments for a given citation i.e. positive, negative, and neutral. Sentiment classification provides us contextual insight into each of the literature citations. Sentiment classification is commonly applied to different domains \cite{bahrainian2013sentiment,wu2015sentiment,feldman2013techniques,lin2009joint,medhat2014sentiment} i.e. movie reviews, product reviews, citations, etc. where a given text string is classified based on its hidden sentiment. Therefore, it is possible to classify sentiments as either subjective \& objective or a more fine-grained classification into positive, neutral, and negative depending on the domain and instances. However, sentiment classification can also induce subjectivity to the opinion.

Sentiment classification provides us a deeper qualitative insight into a given literature citation. However, to get even deeper insights and to evade the likelihood of subjectivity, intent could be identified. The intent of a literature citation refers to the purpose of citing the existing literature. An author can cite a published manuscript for a number of reasons i.e. describing related works, using, extending, or comparing existing approaches and to contradict the claims from previous literature. Intent classification plays a crucial role in validating predicted sentiment of a given citation. The positioning of the citation plays an important role in identifying the intent. For instance, citations usually found in the evaluation and discussion section are more likely to be negative, as the citing authors usually compare the results of their approach in evaluation to prove the superiority of their approach. 

Despite the recently published approaches e.g. Beltagy et al. \cite{beltagy2019scibert} there is still a lack of methods and dataset used for scientific citation analysis. Besides, there is no common definition of intention used to classify publications properly. In this paper, we cleaned the only publicly available dataset for citation sentiment analysis and benchmarked the performance of several models ranging from simple CNN to more sophisticated transformer networks for sentiment and intent classification. By doing so, we achieved a new state-of-the-art for both sentiment and intent classification. We also present the new state-of-the-art as a single solution to be separately trained for sentiment and intent classification. The contributions of this paper are as follows:
\begin{itemize}
    \item We removed the discrepancies and the redundancies present in the previous version of the dataset and made a cleaned and reliable dataset for citation sentiment analysis publicly available\footnote{http://will-be-available-once-accepted/} for the community. 
    \item We conducted performance benchmarking of a set of models ranging from simple CNN based models to sophisticated transformer networks and achieving state-of-the-art performance for both sentiment and intent classification.
    \item We propose one solution for both tasks in hand i.e. sentiment and intent classification. The proposed model can be separately trained for both tasks.
\end{itemize}

\section{Related Work}
In this section, we discuss the existing literature for sentiment and intent classification. We also highlight the key aspects of each existing approaches.

\subsection{Sentiment Classification}
Sentiment classification is a popular task and due to its wide range of applications, there exist numerous publications to address this problem. Tang et al. \cite{tang2014learning} proposed sentiment-specific word embeddings for performing sentiment classification of tweets. Therefore highlighting that the use of highly specialized word embeddings can improve performance for sentiment classification. Thongtan et al. \cite{thongtan-phienthrakul-2019-sentiment} employed document embeddings trained with cosine similarity to perform sentiment classification on a movie review dataset. Cliche \cite{cliche-2017-bb} proposed a sentiment classifier for tweets consisting of an ensemble of CNN and LSTM models trained and finetuned on a large corpus of unlabeled data. 

With the popularity of transformer networks, BERT\cite{devlin2018bert} became a famous choice among the community for a range of Natural Language Processing (NLP) tasks. The BERT model was trained on a large volume of unlabeled data. Therefore, recent literature in the sentiment analysis domain makes use of the BERT model to improve the performance for the task in hand. In \cite{8947435,zhou2016attention,DBLP:journals/corr/abs-1904-12848}, the authors take advantage of transfer learning to adapt pre-trained BERT model for sentiment classification and further boost the performance by complementing it with pre-processing, attention modules, structural features, etc.

The literature discussed so far dealt with sentiment classification in tweets or movie reviews. On the other hand, citation sentiment classification is quite different from review sentiment classification, as the text in scientific publications is formal. Esuli and Sebastiani \cite{esuli2006determining} defined that the sentiment classification is analogous to opinion mining and subjectivity mining. They further discussed that personal preferences and writing style of an author can induce subjectivity in the citations as an author can deliberately make a citation sounding positive or negative.
Athar \cite{athar:2011:SS} performed different experiments using sets of various features like science lexicon, contextual polarity, dependencies, negation, sentence splitting and word-level features to identify an optimal set of features for sentiment classification in scientific publications. Xu et al. \cite{xu2015citation} performed sentiment analysis of citations in clinical trial papers by using textual features like n-grams, sentiment lexicon, and structure information. Sentiment classification is significantly important in the domain of scientific citation analysis due to the scarcity of scientific datasets suitable for scientific sentiment classification and the shallow definition of sentiment for this domain. Finding a sentiment in a text that is written to be analytical and objective is substantially different from doing so in highly subjective text pieces like twitter data.

\subsection{Intent Classification}
The basic concepts of intent classification are the same as sentiment classification. However, contrary to the sentiment classification, the definition of the citation intent classification is much sharper and the label acquisition is strongly related to the sections of a paper where it appears. Usually, section title provides a good understanding of the intent of the citation. However, compound section titles in scientific work can prove to be challenging for identifying the intent. Cohan et al. \cite{cohan2019structural} performed citation intent analysis by employing bi-directional LSTM with attention mechanism and consolidating it with ELMo vectors and structural scaffolds like citation worthiness and section title.

Beltagy et al. \cite{beltagy-etal-2019-scibert} proposed \textit{SciBERT}, which is a variation of BERT optimized for scientific publications and trained on $1.14$ Million scientific publications containing $3.17$ Billion tokens from biomedical and computer science domains. SciBERT was applied to a group of NLP tasks including text classification to sections. Mercier et al. \cite{mercier2019senticite} employed a fusion of Support Vector Machine (SVM) and perceptron based classifier to classify the intent of the citations. They used a set of textual features consisting of type \& length of tokens, capitalization, adjectives, hypernyms, and synonyms. Similarly, Abu-Jabra et al. \cite{abu-jbara-etal-2013-purpose} also employed SVM to perform the intent classification of citations. They suggested that lexical and structural features play a crucial role in identifying the intent of a given citation.

\section{Datasets}
This paper deals with two important aspects concerning citation analysis namely the citation sentiment and intent. For this purpose, we used the following datasets to carry out the evaluations. We identified some inconsistencies in the sentiment dataset, which was later thoroughly cleaned and is being released along with this paper. 

\subsection{SciCite: An Intent Classification Dataset}
In intent classification, we performed our experiments on the SciCite dataset which was proposed by Choan et al.~\cite{cohan2019structural} and covers medical and computer science publications. We chose this dataset for the following reasons: SciCite is a well known publicly available dataset and covers computer science citations. Additionally, it has strong results emphasizing its quality and it is large enough to be used with state-of-the-art deep learning approaches. 

The SciCite dataset has an unbalanced class distribution and consists of coarse-grained labels obtained by clustering citations based on their parent section. According to the authors~\cite{cohan2019structural}, three classes provide a scheme that covers the different intents. Table~\ref{tab:scicite_data} shows the class distribution. The background section provides the majority of citations whereas only a small amount of citations are classified as result or method.

\begin{table}[!t]
\renewcommand{\arraystretch}{1.3}
\caption{SciCite~\cite{cohan2019structural}. Number of instances and class distribution.}
\label{tab:scicite_data}
\centering
\begin{tabular}{|c|c|c|c|}
\hline
 & Class 1: Result & Class 2: Method & Class 3: Background \\
\hline
Train samples & 1109 & 2294 & 4840 \\
\hline
Val samples & 123 & 255 & 538 \\
\hline
Test samples & 259 & 605 & 997 \\
\hline
Overall samples & 1491 & 3154 & 6375 \\
\hline
Class distribution & 13.53\% & 28.62\% & 57.85\% \\
\hline
\end{tabular}
\end{table}

\subsection{CSC: A Citation Sentiment Corpus}
When it comes to the task of citation sentiment classification using publicly available high-quality datasets there is a lack of data. Although, there exist datasets for scientific papers e.g. the dataset proposed by Xu et al.~\cite{xu2015citation} or the sentiment citation corpus proposed by Athar~\cite{athar:2011:SS} these are either not publicly available or have quality issues. Precisely, this problem origins because of the data acquisition and labeling of scientific text as is can not be automated. Conversely, it is straight forward to acquire twitter or movie review data and label it. Due to the lack of other solutions, we had to stick to the dataset proposed by Athar~\cite{athar:2011:SS} although this dataset has a very unbalanced class distribution as shown in Table~\ref{tab:scc_sample_avg}. In the following sections, we refer to this dataset as CSC.

\begin{table}[!t]
\renewcommand{\arraystretch}{1.3}
\caption{Citation sentiment corpus~\cite{athar:2011:SS}. Number of instances and class distribution.}
\label{tab:scc_sample_avg}
\centering
\begin{tabular}{|c|c|c|c|}
\hline
 & Class 1: Positive & Class 2: Negative & Class 3: Neutral \\
\hline
Average Length & 229.4 & 221.8 & 219.6 \\
\hline
Number of samples & 829 & 280 & 7627 \\
\hline
Class distribution & 9.49\% & 3.21\% & 87.30\% \\
\hline
\end{tabular}
\end{table}

\begin{figure}[!t]
\centering
\includegraphics[width=\linewidth]{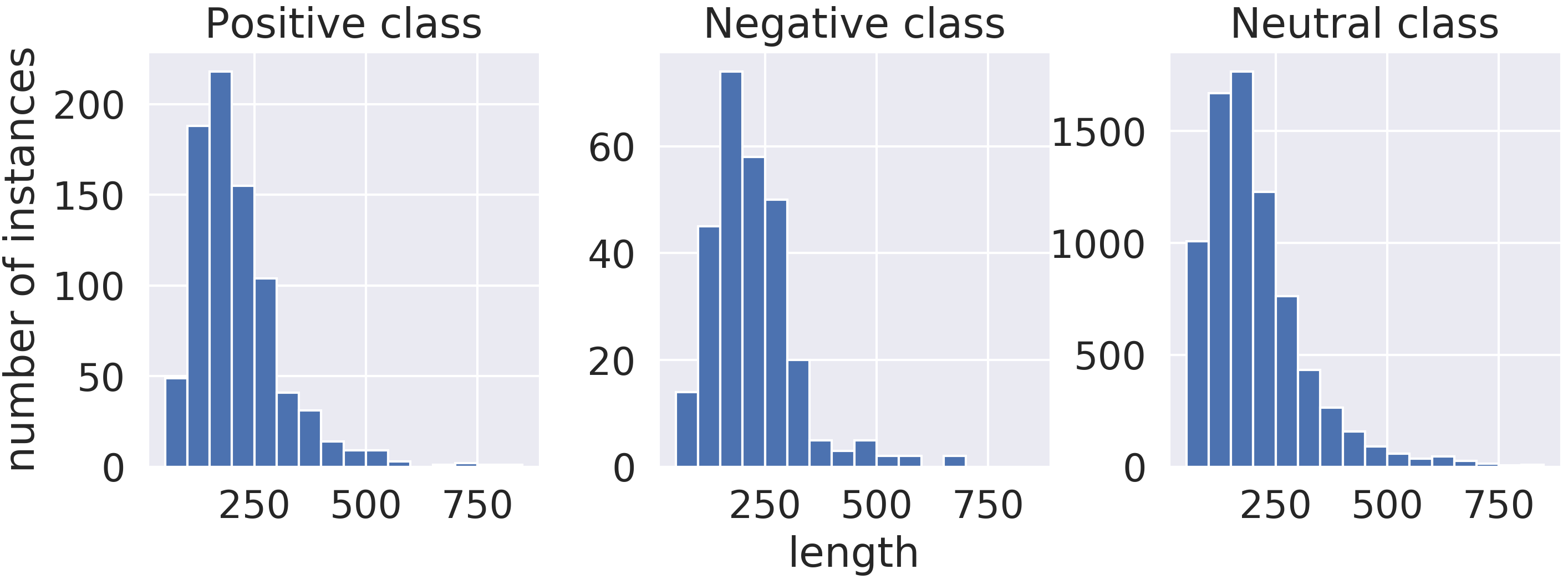}
\caption{Citation sentiment corpus. Sample length class-wise.}
\label{fig:scc_sample_length}
\end{figure}

In Figure~\ref{fig:scc_sample_length} the token length of the samples shows that the sample length is not an indicator for the label. In addition, these numbers demonstrate that a citation contains multiple sentences resulting in an additional context that can be utilized. Extracting only the sentence containing the citation would result in a potential information loss as the sentiment can be included in a follow-up or previous sentence. Therefore, we decided to keep the instances as they are providing us instances of multiple sentences to assure that the content relation can be learned correctly.

\subsection{CSC-Clean: A Cleaned Citation Sentiment Corpus}
During the experimentation phase for this paper we identified several discrepancies concerning duplicated instances, wrong data splits, and samples with impressively bad quality concerning their label consistency. Therefore, it was not possible to compare our approach with the existing results published for the citation sentiment corpus and we decided to clean the dataset to create a novel dataset with better quality covering the same corpus. To do so, we applied the following two steps for dataset cleansing:

\begin{enumerate}
    \item Removing duplicate samples with different labels
    \item Removing duplicate samples with same labels
\end{enumerate}

During dataset cleansing, we removed 756 instances as shown in Table~\ref{tab:scc_comparison}. The removed instances were either identical duplicates of existing instances or provided different labels for the same text. In the case of samples with inconsistent labels, we removed all appearances as a manual selection would induce a bias. We propose the dataset without any duplicates and consistent labels enabling to produce fair and meaningful results using cross-validation to overcome the limited amount of instances for the minority classes. In this paper, we will refer to this dataset as CSC-Clean. The cleaned dataset will be publicly available on the following link: \url{http://will-be-available-once-accepted}. 

\begin{table}[!t]
\renewcommand{\arraystretch}{1.3}
\caption{Comparison of citation sentiment corpus and clean citation sentiment dataset.}
\label{tab:scc_comparison}
\centering
\begin{tabular}{|c|c|c|c|}
\hline
 & Class 1: Positive & Class 2: Negative & Class 3: Neutral \\
\hline
Citation sentiment corpus & 829 & 280 & 7627 \\
\hline
Clean citation sentiment dataset & 728 & 253 & 6999 \\
\hline
Removed instances & 101 & 27 & 628 \\
\hline
\end{tabular}
\end{table}

\section{Experiments and Analysis}
In this section, we will discuss all the benchmarking experiments performed for two different text classification tasks namely citation intent and sentiment classification. We employed models ranging from the baseline models i.e. CNN to highly sophisticated language models i.e. BERT~\cite{devlin2018bert}, ALBERT~\cite{lan2019albert} and XLNet~\cite{yang2019xlnet} based ImpactCite.

\subsection{Intent Classification}

\subsubsection{Experiments:}
For citation intent classification, we performed a bunch of experiments using different models. All the models were trained and evaluated on the SciCite dataset~\cite{cohan2019structural}. We used the train/test split provided with the SciCite dataset and trained three baseline models i.e. CNN, LSTM, and RNN from scratch using a different number of layers, filters, and convolution sizes. In addition, BERT~\cite{devlin2018bert}, ALBERT~\cite{lan2019albert} and ImpactCite used pre-trained model initialization weights and were later finetuned on SciCite dataset. We computed the micro-f1 and macro-f1 as well as the accuracy for each label and each network.

\subsubsection{Results \& Discussion:}
All the evaluation results of citation intent classification are shown in Table \ref{tab:scicite_eval}. it can be observed that the CNN clearly outperformed both the LSTM and RNN. A reason for the worse performance of the RNN is the length of the instances resulting in the vanishing gradient problem, whereas the LSTM processed the citation only in one direction and could not cover the influence on proceeding tokens. We explored different layer and filter sizes for baseline models, however, there is only an insignificant difference when tuning the parameters. Furthermore, CNN was not only superior in performance but also efficient in time complexity than the LSTM and RNN due to the good parallelization. 

The second block of Table \ref{tab:scicite_eval} shows the complex language models whereas the third block shows results from existing literature for citation intent classification. With the fine-tuning of complex language models, we achieved a new state-of-the-art performance by using ImpactCite. ImpactCite significantly outperformed fine-tuned BERT and ALBERT by $3.93\%$ and $4.79\%$ micro-f1 and $5.8\%$ and $6.31\%$ macro-f1 on SciCite dataset. It is to be noted that the accuracy for the classes with less representation in the dataset showed an improvement of about $10\%$, stating that the generalization worked quite well. The performances of our fine-tuned BERT and ALBERT were close to each other showing only an insignificant difference. To conclude, ImpactCite outperformed CNN by $8.71\%$ which highlights the significantly better capabilities of the larger model pre-trained on a different domain and later fine-tuned.

\begin{table}[!t]
\scriptsize
\renewcommand{\arraystretch}{1.3}
\caption{Evaluation results of intent classification on SciCite\cite{cohan2019structural} dataset. L = Layer, F = Filter, C = convolution size.}
\label{tab:scicite_eval}
\centering
\begin{tabular}{|c|c|c|c|c||c|c|}
\hline
\multirow{2}{*}{Topography} & \multirow{2}{*}{Architecture} & \multicolumn{3}{c||}{Class-based accuracy} & \multirow{2}{*}{micro-f1} & \multirow{2}{*}{macro-f1} \\
\cline{3-5}
 &  & Result (\%) & Method (\%) & Background (\%) &  &  \\
\hline
CNN & L 3 F 100 C 3,4,5  & 79.92 & 76.53 & 79.24 & 78.50 & 78.56 \\
\hline
CNN & L 3 F 100 C 2,4,6  & 81.85 & 77.69 & 81.14 & 80.12 & \textbf{80.22} \\
\hline
CNN & L 3 F 100 C 3,3,3  & 64.09 & 71.74 & 85.46 & 78.05 & 73.76 \\
\hline
CNN & L 3 F 100 C 3,5,7  & 76.45 & 74.05 & 85.46 & \textbf{80.49} & 78.65 \\
\hline
CNN & L 3 F 100 C 3,7,9  & 68.34 & 70.58 & 87.26 & 79.20 & 75.39 \\
\hline
LSTM & L 2 F 512 & 73.75 & 73.55 & 79.54 & 76.80 & 75.61 \\
\hline
LSTM & L 4 F 512 & 75.29 & 69.59 & 82.95 & 77.54 & 75.94 \\
\hline
LSTM & L 4 F 1024 & 68.73 & 70.91 & 84.25 & 77.75 & 74.63 \\
\hline
RNN & L 2 F 512 & 25.10 & 56.86 & 62.19 & 55.3 & 48.05 \\
\hline
\hline
ALBERT~\cite{lan2019albert} & Base & 83.78 & 77.03 & 87.06 & 83.34 & 82.62 \\
\hline
BERT~\cite{devlin2018bert} & Base & 84.56 & 75.37 & 89.47 & 84.20 & 83.13 \\
\hline
ImpactCite & Base & 92.67 & 85.79 & 88.34 & \textbf{88.13} & \textbf{88.93} \\
\hline
\hline
BiLSTM-Att~\cite{cohan2019structural} & * & * & * & * & * & 82.60 \\
\hline
Scaffolds~\cite{cohan2019structural} & * & * & * & * & * & 84.00 \\
\hline
BERT~\cite{beltagy2019scibert,devlin2018bert} & Base & * & * & * & * & 84.85 \\
\hline
SciBert~\cite{beltagy2019scibert} & * & * & * & * & * & 85.49 \\
\hline
\end{tabular}
\end{table}

\subsection{Sentiment Classification}
In this section, we will discuss the experiment designs for citation sentiment classification and their evaluations in detail. We adopted a couple of splitting strategies to partition the dataset into training and test set. We performed experiments on the original (CSC) and cleaned the citation sentiment corpus (CSC-Clean).

\subsubsection{Experiment 1: Fixed dataset split on CSC sentiment dataset}
In this experiment we used a fixed $70/30$ training/test split for the existing citation sentiment corpus proposed by Athar~\cite{athar:2011:SS} without any additional data cleaning. This version of the dataset contained the duplicates and inconsistent labels. Similar to citation intent classification, we used three baseline models and three complex language models to perform the experiments for citation sentiment classification. In addition, for the baseline networks, we employed several sample strategies i.e. focal loss, SMOTE \& upsampling, and analyzed their impact concerning the imbalanced data.

\subsubsection{Results and Discussion}
In Table \ref{tab:sentiment_scc} we present the results by using the enhanced baseline approaches as well as three complex language networks. It can be observed that all models were able to capture the concept of neutral citations. Although, the focal loss or SMOTE sampling improved the performance for both the LSTM and the CNN. However, all the models except ImpactCite were unable to classify positive and negative samples with high accuracy. Additionally, we observed that the upsampling method did not improve the performance and rather had a negative impact. ImpactCite showed slightly worse performance on the neutral class, however, it performed significantly better for positive and negative classes. These experiments show that even when used with additional sampling methods the complex language models are superior as they are pre-trained using a large amount of data.

\begin{table}[!t]
\renewcommand{\arraystretch}{1.3}
\caption{Performance: Sentiment citation corpus (CSC)}
\label{tab:sentiment_scc}
\centering
\begin{tabular}{|c|c|c|c|c|}%||c|c|}
\hline
\multirow{2}{*}{Topography} & \multirow{2}{*}{Modification} & \multicolumn{3}{c|}{Class-based accuracy} \\
\cline{3-5}
& & Positive (\%) & Negative (\%) & Neutral (\%) \\ %& micro-f1 & macro-f1 \\
\hline
LSTM & * & 32.8 & 12.4 & 93.9 \\ % & 85.5 & 46.4 \\
\hline
LSTM & Focal & 42.7 & 19.1 & 82.8 \\ %& 76.9 & 48.2 \\
\hline
LSTM & SMOTE & 42.3 & 20.2 & 83.7 \\ %& 77.8 & 48.7 \\
\hline
LSTM & Upsampling & 26.1 & 11.2 & 97.0 \\ %& 87.6 & 44.8 \\
\hline
\hline
RNN & * & 24.5 & 21.3 & 72.7 \\ %& 66.5 & 39.5 \\
\hline
\hline
CNN & * & 28.2 & 21.3 & 94.8 \\ %& 86.2 & 48.1 \\
\hline
CNN & Focal & 36.9 & 16.9 & 94.3 \\ %& 86.4 &\textbf{49.4} \\
\hline
CNN & SMOTE & 39.4 & 20.2 & 84.2 \\ %& 77.9 & 47.9 \\
\hline
CNN & Upsampling & 36.1 & 6.7 & 92.8 \\ %& 84.7 & 45.2 \\
\hline
\hline
BERT~\cite{devlin2018bert} & * & 38.6 & 20.4 & 96.4 \\ %& 88.5 & 51.8 \\
\hline
ALBERT~\cite{lan2019albert} & * & 44.25 & 28.81 & 95.84 \\ %& 88.8 & 56.3 \\
\hline
ImpactCite & * & 78.94 & 85.71 & 75.43 \\ %& 76.1 & \textbf{80.0} \\
\hline
\end{tabular}
\end{table}

\subsubsection{Experiment 2: Cross validation on CSC-Clean sentiment dataset}
In this experiment, we will discuss cross-validation performed on the CSC-Clean. Due to a lack of train/test split of the dataset, Athar \cite{athar:2011:SS} performed 10-fold cross-validation on the original dataset. However, in our case, we performed 10-fold cross-validation on our CSC-Clean. Therefore, we performed ten experiments each using nine out of the ten folds as training and one as test set and averaged their results to compute the overall accuracy. A bunch of experiments was performed employing a variety of models range from baseline CNN models to complex BERT language models.

\subsubsection{Results and Discussion}
In Table~\ref{tab:sentiment_scc_cross} we show the results for the cross-validation of selected models on CSC-Clean. For all baseline models i.e. CNN, RNN, and LSTM, we implemented the class weights to handle the class imbalance problem. Conversely, the results suggest that even after complementing baseline models with elaborated class weights, they are unable to tackle the class weights problem. Therefore, we pre-processed the training set for each fold in which the samples from positive and neutral classes were decreased to the number of negative samples present in the training set of that fold. This pre-processing helped in assuring that each class has equal representation in the training set. It is to be noted that pre-processing was performed on the training set only, whereas keeping the test set intact. 

Additionally, complex language models i.e. BERT, ALBERT \& ImpactCite can effectively fine-tune on small training data as they use their respective pre-trained models. Our results highlight that the baseline-approaches were not able to learn the concept of each class whereas the pre-trained models were able to achieve good results for all classes. As a result, ImpactCite outperformed all other selected models and sets a new state-of-the-art for citation sentiment classification on the CSC-Clean.

\begin{table}[!t]
\renewcommand{\arraystretch}{1.3}
\caption{Cross validation performance: Sentiment citation corpus (CSC-C)}
\label{tab:sentiment_scc_cross}
\centering
\begin{tabular}{|c|c|c|c||c|c|}
\hline
\multirow{2}{*}{Topography} & \multicolumn{3}{c||}{Class-based accuracy} & \multirow{2}{*}{micro-f1} & \multirow{2}{*}{macro-f1}\\
\cline{2-4}
& Positive (\%) & Negative (\%) & Neutral (\%) & &  \\
\hline
LSTM & 34.8 & 19.0 & 92.1 & 84.6 & 46.13 \\
\hline
RNN & 20.7 & 17.9 & 86.0 & 77.9 & 41.53 \\
\hline
CNN & 40.2 & 24.9 & 95.0 & 88.6 & 43,37 \\
\hline
\hline
BERT~\cite{devlin2018bert} & 72.8 & 80.2 & 70.3 & 74.4 & 74.4 \\
\hline
ALBERT~\cite{lan2019albert} & 71.1 & 72.5 & 67.6 & 70.4 & 70.4 \\
\hline
ImpactCite & 64.6 & 86.6 & 82.0 & 77.7 & \textbf{77.73} \\
\hline
\hline
SVM~\cite{athar:2011:SS}\footnote{Trained and tested on CSC} & * & * & * & 89.9 & 76.4 \\
\hline
\end{tabular}
\end{table}

\subsection{ImpactCite: An XLNet-based method for Citation Impact Analysis}
Our evaluation results show that ImpactCite achieved solid results for both the sentiment and intent classification task. ImpactCite was able to handle the long instances and and cover the relation between the sentences within a citation to understand the global context. Conversely, BERT~\cite{devlin2018bert} and ALBERT~\cite{lan2019albert} were not able to do so. However, for the sentiment classification, it is especially important to process the text from both sides to generalize well and deal with the influence of the preceding and following sentences. Additionally, it can utilize the permutations to create synthetic samples to overcome the small amount of data provided for the sentiment task. Therefore, ImpactCite achieved a state-of-the-art performance for both tasks. We propose ImpactCite, an ImpactCite-based solution covering both the sentiment and intent classification which leads to a qualitative citation analysis.

\section{Conclusion}
Our comprehensive experiments show the improvements in both the sentiment and the intent classification task for citations in scientific publications encouraging the use of those two properties to provide better information about the influence of papers. Also, we achieved state-of-the-art performance for the intent classification on SciCite \cite{cohan2019structural} dataset and sentiment classification on our novel citation sentiment dataset. Our results increased the SOTA for SciCite to $88.93\%$ using ImpactCite which is an increase of $3.44\%$ compared to SciBERT. Furthermore, for the sentiment citation corpus, we pushed the old state-of-the-art result of $76.4\%$ to $77.73\%$. Also, we compared the results for the different classes to highlight that the performance for two out of the three classes improved significantly. Our study emphasizes that recent transformer-based and auto-regressive models are far superior compared to simpler approaches like LSTM or CNN. Concerning the sentiment classification, we emphasize that the ImpactCite is much more robust for small or large datasets with long sequences and significantly outperforms other existing state-of-the-art methods. 
 \bibliographystyle{splncs04}
 \bibliography{bibliography}
\end{document}